\RequirePackage{fix-cm}
\documentclass[smallextended]{svjour3}       
\smartqed  
\usepackage{graphicx}
\usepackage{textgreek}
\usepackage{algorithmic}
\usepackage{booktabs}
\begin{document}

\title{An Interpretable Machine Learning Framework for Non-Small Cell Lung Cancer Drug Response Analysis
}
\subtitle{}


\author{Ann Rachel \and Pranav M Pawar \and Mithun Mukharjee \and Raja M \and Tojo Mathew
}


\institute{Ann Rachel, Pranav M Pawar, Mithun Mukherjee, Raja M, Tojo Mathew \at
              Department of Computer Science\\
              Birla Institute of Technology and Science Pilani, Dubai Campus, Dubai, UAE\\
              \email{f0220078@dubai.bits-pilani.ac.in, pranav@dubai.bits-pilani.ac.in, mithun@dubai.bits-pilani.ac.in, raja.m@dubai.bits-pilani.ac.in, tojomathew@dubai.bits-pilani.ac.in}           
}

\date{Received: date / Accepted: date}

\maketitle

\begin{abstract}
Lung cancer is a condition where there is abnormal
growth of malignant cells that spread in an uncontrollable fashion
in the lungs. Some common treatment strategies are surgery,
chemotherapy, and radiation which aren’t the best options due
to the heterogeneous nature of cancer. In personalized medicine,
treatments are tailored according to the individual’s genetic in-
formation along with lifestyle aspects. In addition, AI-based deep
learning methods can analyze large sets of data to find early signs
of cancer, types of tumor, and prospects of treatment. The paper
focuses on the development of personalized treatment plans using
specific patient data focusing primarily on the genetic profile.
Multi-Omics data from Genomics of Drug Sensitivity in Cancer
have been used to build a predictive model along with machine
learning techniques. The value of the target variable, LN-IC50,
determines how sensitive or resistive a drug is. An XGBoost
regressor is utilized to predict the drug response focusing on
molecular and cellular features extracted from cancer datasets.
Cross-validation and Randomized Search are performed for
hyperparameter tuning to further optimize the model’s predictive
performance. For explanation purposes, SHAP (SHapley Additive
exPlanations) was used. SHAP values measure each feature’s
impact on an individual prediction. Furthermore, interpreting
feature relationships was performed using DeepSeek, a large
language model trained to verify the biological validity of the
features. Contextual explanations regarding the most important
genes or pathways were provided by DeepSeek alongside the top
SHAP value constituents, supporting the predictability of the
model .
\end{abstract}

\section{Introduction}
\label{intro}
There has been a revolutionary shift in modern on cology with the emergence of personalized medicine, which provides therapeutic strategies based on phenotypic and genotypic patient data. According the the World Health Organization (WHO), one of the leading causes of cancer related deaths around the world is lung cancer [1]. Due to the significance of lung cancer, such an approach, that is, using personalized medicine along with artificial intelligence can make a significant difference in the field of oncology. Two of the predominant subtypes of Non-small cell lung cancer are: lung adenocarcinoma (LUAD), and lung squamous cell carcinoma (LUSC), which comprises of approximately 85\% of all lung cancer cases [2]. Both these types exhibit different molecular profiles and response to therapy, even though they originate from the same location, in the lung epithelium. Cancer’s heterogeneity makes it quite difficult to treat it and attain optimal outcomes, which is why there is a need to develop models based on large data that can predict how a drug will respond for a specific patient.
Precision oncology can be enabled using the powerful tool, machine learning, which helps integrate and analyze biomedical data. This paper focuses on LUAD and LUSC type of cancer, where AI models can be trained on multi omics datasets to predict the sensitivity of cancer cells to various drugs. With the help of AI, hidden patterns in complex datasets can be identified which are generally difficult to detect using normal statistical approaches
\subsection{Traditional Methods}
Cancer or malignant neoplasm, a genetic condition, is brought on by genetic or epigenetic changes in somatic cells. Radiotherapy, chemotherapy, and surgery are the three main treatments for cancer [3]. Each treatment approach has drawbacks and related issues, including myelosuppression, symptoms in the gastrointestinal tract, toxicity to the liver and kidneys, and damage to the heart. Because of the heterogeneity of tumors and the emergence of resistance, pharmaceutical treatments are ineffective. Surgery frequently lowers quality of life by increasing mortality and morbidity. Historically, generic, "one size fits all" methods have been used to treat cancer [4]. The effectiveness of these treatments varies greatly from person to person, and they frequently damage healthy, noncancerous organs and tissues. To increase patient survival and quality of life without sacrificing therapeutic efficacy, alternative therapies are required. 
 
\subsection{Personalized Medicine in Lung Cancer}
 Personalized medicine helps facilitate the early diagnosis, treatment, and prevention of a disease by taking into account the patient’s genetic data, environmental factors, and lifestyle decisions. By considering an individual’s genetic profile, it increases the potential to predict which medical treatments are safer and effective for them while at the same time minimizing adverse reactions[5]. This paradigm has brought about a more individualized approach with regard to the patient rather than the generic treatment approach. The main objective of the paper is to propose an exhaustive model capable of efficiently analysing large-scale quantities of healthcare data, uncovering trends and insights for the enhancement of patient-oriented individualized therapy. In addition to precision medicine, the use of artificial intelligence in healthcare has the potential to change aspects such as patient monitoring, clinical decision-making, and drug development and, therefore, accelerate medical progress.
 
 One field of application is oncology, as the heterogeneity of cancer makes it ineffective to use standard approaches such as chemotherapy and radiation, as these are effective only in a fraction of patients. In the field of personalised medicines with respect to cancer, AI can help in cancer detection and classification, drug discovery and repurposing, and patient treatment outcome prediction [6]. With the help of deep learning techniques, molecular profiling and specific mutations analysis, models trained on genomic data can help predict drug responses specific to that patient facilitating targeted cancer therapy.
Additionally, AI based deep learning methods can analyse large sets of data to find out early signs of cancer, types of tumors, and prospects of treatment.
\begin{figure}
    \centering
    \includegraphics[width=1\linewidth]{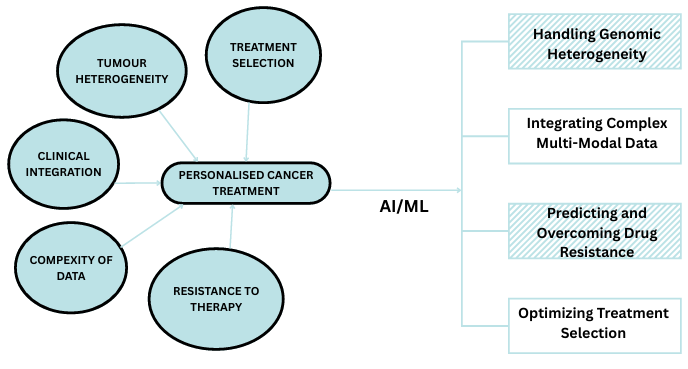}
    \caption{AI/ML for Personalised Cancer Treatment}
    \label{fig:placeholder}
\end{figure}

\subsection{AI Applications}
 The integration of artificial intelligence with personalized medicine can help to design better treatment plans for each individual. Suspicious lung nodules can be identified as benign or malignant using AI, as it can analyze CT, MRI or
 PET scans and help in earlier diagnosis and treatment. Also, based on imaging and clinical data, machine learning models can predict the overall survival rate. AI can also help identify mutations and determine how well a patient may react to a drug or certain treatment. Some common AI techniques used in this field are, machine learning and deep learning to predict disease risk, classify the cancer subtypes and understand medical images[7]. Natural Language Processing can be used to extract information from Electronic Health Records (EHRs) and identify adverse drug reactions. Reinforcement Learning can help optimize dosing schedules and create personalised treatment plan.
 
 This work mainly focuses on Drug Response Prediction where genetic and clinical data are being analysed using AI techniques to identify the most effective drug for a lung cancer patient while at the same time minimize any adverse effects. The model is further improved to predict drug sensitivity based on clinical data like gene expression, mutations and other biomarkers[8]. The model also implements explainable techniques like SHapley Additive exPlanations (SHAP) to provide insights on how each feature affects the target variable(LN-IC50) and by how much. We attain a more contextual understanding by passing SHAP values into DeepSeek which provides more information about the drug’s role, its metabolism, treatment adjustments and actionable steps to be taken by the clinician. By integration of AI with personalized medicines, the paper aims to improve precision oncology.

\subsection{\ Key Components of the Proposed Method}
The paper is focused on the development of personalized treatment plans using specific patient data, including medical history and genetic profile. To achieve actionable and precise outcomes there are several key steps are followed:

\begin{enumerate}
    \item \textbf{Data Pre-processing:} Missing and inconsistent values are handled such that the dataset is made ready for model training. One-hot  encoding is performed to handle categorical values. The most relevant variables for prediction are obtained through feature selection process.

    \item \textbf{Model Selection and Training:} Personalized therapies are delivered using the patient-specific profile and employing machine learning algorithms like XGBoost to predict sensitivity of any drug to an individual based on their genetic profile. Hyperparameter tuning is performed using RandomizedSearchCV to identify the best parameters to use, enhancing the model's performance.

    \item \textbf{Validation and Optimization:} Accuracy and reliability in predictions are improved using hyperparameter optimization. The models are then tested with cross-validation.

    \item \textbf{Model Interpretability:} To understand the model and its results, interpretability is applied. Explainability techniques like SHAP and feature importance analysis are employed to understand how specific features influence the prediction. SHAP values are also passed to DeepSeek, a large language model, to interpret feature annotations and extract biological significance from literature, helping us understand why certain genomic or drug features are influential.
\end{enumerate}

 Through our comprehensive analysis of drug sensitivity focusing on lung cancer, we are able to provide insights on the various factors that affect the sensitivity of drugs to improve decision making in the field on oncology. This article contains a study about the research done in precision medicine focusing on drug sensitivity prediction and the problems addressed by various research papers. With the implementation of various machine learning techniques along with genetic data, we made a significant improvement for cancer patients by focusing on their individualised genetic data. 

\section{Related Works}

\subsection{ML/DL Applications for Personalised Medicine in Cancer}

The use of machine learning models have been discussed in [9] to distinguish cancer and normal samples from genomic data with high accuracy based on classifiers trained on 19,627 genes from whole genome sequencing (WGS) data. Based on data from The Cancer Genome Atlas (TCGA) and Genotype-Tissue Expression (GTEx) projects, the models are highly precise, sensitive, and specific and are promising for early cancer detection and diagnosis. However, one of the major limitations of this approach is sole reliance upon genomic data that cannot fully address the complexity of cancer biology. The study has not addressed combining other omics data—i.e., transcriptomics, proteomics, and metabolomics, which could possibly better guide the predictions and unravel more holistic tumor biology. This lack of multi-omics integration can limit diagnostic precision, hinder the identification of new biomarkers essential to early diagnosis and treatment, and reduce the model's effectiveness in personalized medicine through the neglect of individual variability in tumor biology. Bridging this research gap might greatly improve models of cancer classification, offering a more complete and precise approach to cancer diagnosis and treatment.

\vspace{1em}
The article [10] aims to improve cancer patient stratification by leveraging both phenotypic and genetic features obtained from electronic health records and genetic test reports, appreciating the potential of uniting these information sources for more precise treatment choice. The method relies on a three-step process of feature pre-processing, classification of cancer patients, and clustering based on machine learning models such as Logistic Regression (LR), Multi-Layer Perceptron, Random Forest, and Support Vector Machine. Node2vec embeddings are also used in the research to enhance association representation between phenotypic and genetic features. The database, compiled from Mayo Clinic's electronic health records and genetic reports, facilitates full analysis since the results indicate joint phenotypic and genetic features that improve the accuracy of classification much more than either dataset alone. However, the study also indicates strong limitations in the shape of a small size for the dataset that affects the generalizability of the model and variation in performance with classifiers. Interestingly, though convolutional neural networks were experimented with, they fared poorly compared to other models with an F1 score of 0.74, suggesting the inability to handle the complexity and high dimensionality of genomic information.

\vspace{1em}
This paper [11] focuses on the use of AI in cancer research and precision medicine and discusses how its potential can improve cancer care by leveraging advanced data analysis and machine learning modeling. These include molecular characterization deep learning models that analyze multidimensional data to detect, classify, and predict the treatment outcomes of cancer. Three key genomic data sets are employed, i.e., TCGA, PCAWG, and METABRIC, but the study shows that these data sets hold attributes bias, especially race and ethnicity. The main strength of AI is how it handles intricate data sets in a quick time, which results in better identification of cancer cases and formulation of personalized treatment plans considering genetic and environmental factors. The limitations are dataset biases that reduce model generalizability, and dependence on loosely structured data, which adds illogical inconsistencies to the models. The gaps that these biases leave are in regions of the lack of rich datasets that cover the entire population and increased linkage of information across disparate sources. For the purpose of accuracy, the primary support pillars of concern pertain to model verification based on several independent datasets with different 0.8 strong AUC scores, which is clinically distinct. To address the problem of overfitting, strategies involving cross-validation, increase in the size of the training set, and the use of ensemble methods are advised to increase the stability of the model. While Electronic Health Records (EHRs) provide valuable patient data, they are yet to be optimally used in AI models as to its unstructured nature which requires massive curation. Furthermore real-time evaluation to predict cancer risk can be enabled by the integration of genetic susceptibility, EHRs, and lifestyle thereby providing personalized intervention and risk management.

\vspace{1em}
The article [12] attempts to harmonize genetic and phenotypic data to make primary cancer type predictions and predict as-yet unknown primaries, permitting earlier diagnosis and optimal treatment selection. The study employed machine learning and deep learning-based algorithms, including a stacked sparse auto-encoder-based classifier and artificial neural network tuned using genetic algorithm, that were trained using combined electronic health record and genetic data. The information integrates genetic data from oncology reports and EHRs and provides a full analysis that improves the accuracy of predictions, with high Area under the Receiver Operating Characteristic Curve (AUROC) scores (e.g., 99.5\% for lung cancer) and clinical validity guaranteed through real-patient validation. Some of the limitations are the small dataset size, which affects robustness, and genetic data bias because germline and somatic mutations cannot be distinguished. The study's research gaps are for larger, more diverse datasets for validation and the challenge of representing genetic data with relational databases, particularly anonymization and aggregation. The method was robust in predictive accuracy with a mean AUROC of 96.56\% in classifying primary cancers and 80.77\% in predicting unknown primaries. This research overcomes previous overfitting and less-than-optimal utilization of EHRs through leveraging a network-structured data representation using HL7 FHIR [13] standards and Node2vec embeddings for improved cancer prediction. An important challenge remains: the genetic reports only considered somatic mutations, making it impossible to distinguish from germline mutations, thereby introducing bias and weakening the findings.\vspace{1em}

This work by D' Amico et al. [14] discusses the AI-generated synthetic data's intended role in advancing research and precision medicine in hematology. This paper will focus on the creation of synthetic datasets for hematological neoplasms while ensuring fidelity and privacy using a Synthetic Validation Framework (SVF). The article attempts to address the problem of improving the integrity and privacy of synthetic data so that the data might be used most efficiently in clinical trials and translational research. This is because improper synthetic data may compromise any research conclusions drawn and patient privacy. Synthetic data aims to solve data imbalance, increases dataset, and keeps privacy intact; however, challenges remain in fidelity and possible misuse. Key knowledge gaps include clinical validation and longitudinal applicability.
That supported accuracy was determined based upon the Clinical Synthetic Fidelity (CSF) and Genetic Synthetic Fidelity (GSF) metrics. Clinical synthetic fidelity was found to be 90\%for acute myeloid leukemia, while genomic synthetic fidelity reached 88\%. For a wider International Working Group for Prognosis in Myelodysplastic Syndromes (IWG-PM) Myelodysplastic Syndromes (MDS) cohort, CSF and GSF reported mainly at 93\%. Thus, the clinical validation above determines reliability across clinical contexts. To mitigate dataset limitations, conditional Generative Adversarial Network is used in generating synthetic data and performing data augmentation, which enhances the robustness of the model. By validating synthetic cohorts in multiple patient datasets, this research ensures their clinical relevance, placing balanced focus and priority on fidelity and privacy for an effective role in clinical trials and translational research. 

With a focus on the integration of multi-omics data, including genomic, metabolomics, and imaging data, to improve disease knowledge and patient-specific treatment methods, this paper [15] describes how machine learning techniques can be utilized in personalized medicine to handle complex data. Machine learning methods helps in seeing patterns that conventional statistical techniques might miss, enabling more precise classifications and predictions. Citing earlier research that used genetic, imaging, and clinical data for improved illness prediction and categorization, the paper also addresses practical uses of multi-omics integration.  Research studies are also illustrated where the work is always being advanced in integrating several parameters from biology and the environment in machine learning environments. Methodologically, the paper also discusses various machine learning frameworks, including convolutional neural networks and Bayesian networks, applied to real data in order to derive useful insights. However, one significant disadvantage of the research is that it does not include external validation in the real world, as it does not adequately test these machine learning models using external clinical datasets. Also, although it includes many different machine learning approaches, it is devoid of an exhaustive study of deep learning methods, which have proved to show excellent outcomes in dealing with multi-modal biomedical data. To bridge these gaps would further enhance the contributions of the study by establishing the real-world applicability of the suggested methodologies in clinical procedures. 

The paper [16] discusses improvements in pancreatic adenocarcinoma grading via Bayesian Convolutional Neural Networks for better prediction accuracy, incorporating uncertainty into the model's action, which is imperative in the prevention of clinical misdiagnosis. By exploiting transfer learning with architecture like DenseNet-201, VGG-19, and ResNet-152V2, the model is tuned on a relatively small dataset containing 3201 labeled high-resolution tissue sample patches across different cancer grades. Uncertainty quantification allows clinicians to rank cases for further review, whereas transfer learning helps out with generalizing where large datasets are not available. This introduces learning difficulties when estimating the posterior distribution because of computational budgets, and although the impact of class imbalance is removed, it remains a quality challenge in some datasets. However, a research gap was identified concerning the relationship between uncertainty and accuracy; such would need to increase the acceptance threshold of clinical predictions. The main factor is thus accuracy, since malpractices in pan cancer misdiagnosis have dire consequences, and the study found a great correlation (ρ > 0.95) between uncertainty and prediction errors, and therefore pitched the idea of uncertainty-aware metrics like ARQ that can balance classification accuracy and misclassification risk. The study also attempted to address issues regarding previous variant misclassification by making use of uncertainty estimation, allowing for a reject option against uncertain predictions, and improving model responses to out of distribution samples. However, a very key limitation remains, a small dataset size, which might inhibit generalizability despite transfer learning. The dataset construct, where training, validation, and testing accounted for 60\%, 20\%, and 20\% respectively, means the model is restricted to little variability in cases, thus presenting a chance of overfitting, something that undermines its robustness for application in clinical settings.
\subsection{\ AI Applications of Personalised Medicine in Lung Cancer}
\begin{table*}[!htbp]
  \centering
  \caption{Summary of studies applying ML/DL to cancer classification and prediction using heterogeneous data sources.}
  \resizebox{0.95\textwidth}{!}{
    \begin{tabular}{|p{2.2cm}|p{3.4cm}|p{3.1cm}|p{2.5cm}|p{3.1cm}|}
      \hline
      \textbf{Reference} & \textbf{Contribution} & \textbf{ML/DL Method} & \textbf{Cancer Type/Focus} & \textbf{Result} \\ \hline
      [9] A. Hooshmand (2020)& Proposed a machine learning system to classify genomic profiles for accurate cancer diagnosis. 
      & Naïve Bayes, SVM, Decision Trees, Random Forest, Logistic Regression, K-NN.
      & 22 cancer types using 19,627 genes from WGS data.
      & Reported precision, sensitivity, and specificity of 1.0 for nearly all cancer types, indicating excellent classification performance. \\ \hline

      [10] D. Oniani et al. (2021)
      & Developed a classification framework combining phenotypic and genetic features from EHRs using multiple GNN models.
      & Eight different Graph Neural Network architectures explored.
      & Multiple cancer types, with focus on phenotypic-genotypic interactions.
      & CNN achieved F1-score of 0.74; Logistic Regression scored 0.90. \\ \hline

      [11] B. Bhinder et al. (2021)
      & Demonstrated AI's role in risk assessment and personalized care by integrating diverse omics and clinical datasets.
      & Deep Neural Networks, Random Forest, DeepVariant.
      & CNS and skin cancers; includes methylation and mutation analyses.
      & No specific metrics reported; emphasized AI's utility and lab-level validation. \\ \hline

      [12] N. Zong et al. (2021)
      & Integrated genetic reports and EHRs to predict and classify both primary and unknown cancer types.
      & Node2vec embeddings, CNNs, Autoencoders.
      & Multiple cancers incl. prostate, breast, liver, pancreas, and thyroid.
      & Achieved 95\% AUROC for primary classification and 80\% for unknown primaries. \\ \hline

      [14] S. D'Amico et al.(2022)
      & Investigated AI-generated synthetic data as a privacy-preserving solution for accelerating hematological cancer research.
      & Conditional GANs (cGANs) for synthetic clinical-genomic data generation.
      & Hematological malignancies: Myelodysplastic Syndromes, AML.
      & Used NNDR (Nearest Neighbor Distance Ratio) between 0.60-0.85 to balance fidelity and privacy. \\ \hline
      [15] S. J. MacEachern and N. D. Forkert (2021)
      & Integrates different types of data, known as multi-modal or multi-omics data, to increase the level of precision medicine. 
      & Deep learning algorithm, such as DeepSEA and DeepVariant, that is applied in genomic data and in estimating the functional effects of genetic variants.
      & The paper mentions the application of machine learning in various cancer types, not focusing on a certain type.
      & It highlights that verification of the models through independent test sets is an important factor to take care of, so the model does not do overfitting or underfitting of the data. \\ \hline
      [16] B. Ghoshal and A. Tucker (2022)
      & This paper utilizes uncertainty in the automated assessment of pancreatic adenocarcinoma through histopathology images. 
      & It uses an algorithm grounded in Bayesian Convolutional Neural Networks capable of estimating uncertainty associated with predictions made by the model.
      & The dataset investigated within this research comprises histopathology images focusing on pancreatic adenocarcinoma.
      & The paper elaborates further on the merit of metrics that apply relative weights to classification accuracies against misclassification costs.
      \\ \hline
      [17] A. Rafferty et al. (2024)
      & The paper proposes an interpretable AI model for lung cancer detection using concept bottleneck models to enhance transparency.
      & InceptionV3 for concept prediction; MLP, SVM, and DT for label prediction.
      & Lung cancer detection using chest X-ray images.
      & Achieved F1-score > 0.9 for lung cancer detection, showing strong classification performance. \\ \hline

      [18]Y. Jiang et al. (2024)
      & Proposes WRE-XGBoost algorithm to enhance prediction of anti-cancer drug sensitivity and tailor treatment plans.
      & WRE-XGBoost (an optimized variant of XGBoost with weighted feature selection).
      & Multiple cancer types; focus on genes modulating chemotherapy response.
      & No specific metrics reported; emphasizes improved prediction through refined feature selection. \\ \hline

      [19]R. Qureshi et al. (2022)
      & Focuses on developing a personalized drug response prediction model tailored to the needs of lung cancer patients.
      & The best performing model was constituted of an XGBoost classifier.
      & It's primary focus was lung cancer, particularly patients with mutations in the epidermal growth factor receptor gene.
      & The PDRP model demonstrated an unprecedented accuracy in predicting the levels of response to prescribed drugs.\\ \hline
      [20]J. R. Astley et al. (2024)
      & Sharpens survival prediction in patients with non-small cell lung cancer (NSCLC) undergoing radical radiotherapy.
      & Utilizes Cox Proportional Hazards model, random survival forests, and deep learning models to predict overall survival from pre-treatment covariables.
      & Predicting overall survival in non-small cell lung cancer patients.
      & The models' performance was primarily expressed in terms of C-index and integrated Brier score. The DL method promised a C-index of 0.670 and an IBS of 0.121. \\ \hline
      [21]L. Pant et al. (2024)
      & Leverages genomic data to predict drug sensitivity in cancer and improve treatment outcomes by providing patients with tailored therapies.
      & The authors use several machine learning methods to analyze complex genomic data and predict drug response.
      & The study examines cancer cell lines to assess how sensitive these cell lines are to a range of anti-cancer drugs.
      & The findings show Random Forest was better than the other models in predicting drug sensitivity.\\ \hline

      [22]Y. Shi et al. (2024)
      & The paper provides a full, searchable database (D3EGFRdb) of patient cases with EGFR mutations and treatment responses.
      & The study utilizes a deep learning model to predict drug sensitivity in patients with EGFR mutations, employing underlying neural network architectures .
      & The study focuses on non-small-cell lung cancer and the association of EGFR mutations on drug sensitivity.
      & The results show that the D3EGFRAI model was satisfactorily predictive of drug sensitivity, showing potential utility for clinical decision making.\\ \hline

    \end{tabular}
  }
  
  \label{tab:full_width}
\end{table*}

The work demonstrated in this paper[17] is aimed at developing a transparent and clinically interpretable AI system for lung cancer detection in chest X-rays through an ante-hoc approach that relies on concept bottleneck models. Through this approach, everything from the AI's decision-making process is transparent for enhanced insight, solving the limitations of conventional post-hoc techniques, such as LIME or SHAP which typically establish unreliable explanations. Tested on 2,374 scans obtained from the MIMIC-CXR ((Medical Information Mart for Intensive Care – Chest X-ray) )dataset, the proposed model achieved an F1-score above 0.9 and a concept precision of 97.1\%, significantly outperforming other existing XAI tools, such as CXR-LLaVA (Chest X-ray Large Language and Vision Assistant ). While this study improves classification performance and generates clinically related explanations, such factors' reliability may vary based on dataset quality and the threat of false negatives. The study indicates a gap in the existing techniques of XAI and offers the development of AI systems that embrace clinical concepts into the real decision-making process, for improved trust and applicability in medical setups. Addressing the question of explanation reliability, this paper puts forward a promising pathway, would enhance the introduction of AI into lung cancer diagnostics, improving both the fidelity of models and clinicians' trust.\vspace{1em}

This paper by Jiang et al. [18] proposes a novel approach that aims to predict anti-cancer drug sensitivity using a new algorithm, WRE-XGBoost, which integrates weighted feature selection to increase accuracy. The WRE-XGBoost algorithm uses a machine learning approach to directly improve feature relevance for predictions pertaining to drug sensitivity. This approach enables a better focus on data analytics which can significantly improve predicting the response of cancer patients to various treatment options. The WRE-XGBoost algorithm is an extension of XGBoost and WRE-XGBoost, the latter two are well-known for their efficacy and effectiveness in dealing with vast amounts of information and intricate feature relationships. Weighted Relevance Estimation  (WRE) indicates a weighting mechanism that selects the most impactful features in the WRE-XGBoost dataset. The WRE-XGBoost algorithm's major advantage is its focus on relevant feature selection which improves accuracy to enhance personalized adaptive cancer treatment. One disadvantage of this design may be the detailed algorithm in design that requires extensive skill, cost, and computing power to implement realistically making it less available for widespread use in clinical setting.\vspace{1em}

The focus of the article [19] was creating a machine learning model, which they named Personalized Drug Response Prediction (PDRP), that would predict drug response levels in lung cancer patients by integrating clinical, demographic, energetic, and geometrical features. Performing Molecular Dynamics (MD) simulations to model the binding site of drug-target complexes. It involves using machine learning classifiers and focusing on the XGBoost classifier for data analysis concerning drug responses to be predicted. The dataset incorporated demographic and clinical data for lung cancer patients, which included: age, gender, history of smoking, survival status, and level of response to treatment. A total of 33 different Epidermal Growth Factor Receptor (EGFR) point mutations were considered, with molecular dynamics (MD) simulations conducted for every mutant. The primary algorithm performed was the XGBoost classifier, renowned for exemplary performance in classification problems. The model was conditioned on the features to ascertain the four stratified responses: complete response, partial response, stable disease, or progressive disease. An overarching benefit of the PDRP model is the precision rate it has in accuracy known to be 97.5 percent when spot on classified. A noted drawback was the male dataset from which misclassifications were drawn.The efficiency of the PDRP model was assessed in terms of its precision, recall, F1-score, and balanced accuracy. The model exhibited a classification accuracy of 97.5 percent, which is an indicator of its productivity in predicting the responses to drugs.\vspace{1em}

The objective of the study [20] by J. R. Astley et al.[] is to generate the OS of patients with NSCLC receiving radical radiotherapy by applying different modeling techniques to boost predictive accuracy. The Deep Learning model performed prominently among others, yielding a C-index of 0.670 and an IBS of 0.121. The authors also introduce Local Interpretable Model-agnostic Explanations (LIME), which were vaguely addressed in past research, in order to enhance the interpretability of the model regarding feature contributions. While this advancement renders the model better suited for clinical applications, the study acknowledges limitations such as:
•	LIME reliability regarding binary features: The perturbation-based approach of LIME with binary variables may fail to accurately grasp the importance of binary covariables.
•	Discrepancy between local vs. global interpretability: LIME provides local explanations which do not necessarily correlate with a global understanding of the model behavior.
To address these concerns, the authors propose having SHAP values incorporated along with LIME for well-rounded explanations. Furthering, although the dataset consisting of 471 NSCLC patients receiving radical radiotherapy has a good basis for survival prediction, the small number of stage IV patients could severely limit generalizability. The study raises a critical gap within research in interpretable tools in machine learning-based survival prediction along with further needs for validation. The future will be focused on embedding local and global explainers to improve transparency and confidence in clinical decision-making based on LIME and SHAP values. \vspace{1em}

The objective of this research paper  [21] is to develop personalized medicine for cancer treatment through the use of genomic data to determine drug sensitivity. The ultimate goal is to improve the patient-specific therapeutic interventions delivered to patients by way of the unique genetic and molecular profiles of individual tumors, rather than the existing standard of care that is delivered uniformly to all patients. In order to develop the patient-specific therapeutic interventions, the data generation involved feature engineering and feature selection, which is the transformation of raw genomic data to be input into machine learning models. This involved processing the genomic data, feature engineering, and finally feature extraction which allowed for the complex genomic data to be summarized into reasonable metrics. This last stage included dimension reduction methodologies, for example using Principal Component Analysis (PCA) and autoencoders to maintain informative patterns while reducing the number of original variables. This study uses the Genomics of Drug Sensitivity in Cancer (GDSC) dataset, which provides an extensive genomic profile for cancer cell lines along with sensitivity to a large range of anti-cancer drugs. The GDSC dataset is significant because it provides the necessary details regarding the biological variability in response to drug sensitivity regarding genomic alterations. The study utilizes many machine learning algorithms, for example using Logistic Regression as an initial model for binary classification of drug sensitivity (sensitive or resistant), Random Forest as an ensemble option that fits many decision trees to predict outcomes correctly and to improve accuracy, and finally using XG-Boost which is a hyper-optimized version of a gradient boosting models and improves interpretability from phasing structural info together while decreasing training time and improving prediction performance.The positive aspects of this approach included increased accuracy for predicting drug sensitivity with the use of genomic somatic data and the potential for individualized treatments leading to better treatment outcomes and decreased toxicity to patients. However, the study included some negative aspects as well, particularly related to utilizing highly dimensional genomic data which created complexity for training models and model interpretation, and also potentially causing logistic regression to struggle with high-dimensional datasets where the number of features exceeded the number of samples. Overall, the random forest method showed better overall performance than the other modeling methods and in the various metrics of measuring prediction risk for drug sensitivity. The study expressed the objective of improving personalized cancer treatment through genomic data and advancing data analysis on genomic molecular cancer data by using machine learning algorithms to maximize predictive accuracy while acknowledging the challenges associated with using complex datasets.\vspace{1em}

The main goal of the study [22] was to assess the effect of Epidermal Growth Factor Receptor (EGFR) mutations on drug sensitivity and how to best provide treatment with a database of actual patient cases and a drug sensitivity prediction tool. The study involved the integration of a clinical patient database (D3EGFRdb) and a drug response prediction model (D3EGFRAI). The database contains patient cases with EGFR mutations, along with clinicopathological characteristics and instrumental therapies in terms of approved drug responses. The model was developed by deep learning to evaluate drug sensitivity. The external clinical dataset included clinical records and outcomes for 102 patients treated with EGFR-Tyrosine Kinase Inhibitors (EGFR-TKIs) at the Shanghai Pulmonary Hospital from March 2015 to October 2020. The dataset included information pairs of various drugs and mutants, and was complied according to Response Evaluation Criteria in Solid Tumors v1.1. The study employed a deep learning model for prediction of drug detected and used drug and protein encoders from DeepPurpose. The model was pretrained in a large scale bioactivity dataset, where a fraction of the dataset was used as a test set and the remainder divided into the training and validation set. One advantage of the D3EGFR platform is that it streamlines with real patient cases that feature specific clinical information and medication outcomes, while also allowing patients to be reasonably correct on treatment.Moreover, the prediction model produced acceptable results in clinical patient cases, and therefore, increasing the possibility of precision medicine. A potential disadvantage argued in the study is how the study is reliant on retrospective data, as this could introduce systematic biases and constrain the generalization of the findings. In addition, the prediction models accuracy and reliability could be improved with more reported and internal clinical trial results in the future. While certain performance measure values (like accuracy, sensitivity, or specificity) were not provided in the contextual descriptions, the study states the D3EGFRAI model has produced satisfactory prediction performance for clinical patient cases, indicating a good evaluation of its effectiveness.\vspace{1em}

\section{Methodology}
This section provides the methodology adopted in this work for predicting drug sensitivity in lung cancer patients for two subtypes of non-small cell lung cancer, i.e., lung adenocarcinoma (LUAD), and lung squamous cell carcinoma (LUSC) with the help of explainable AI and machine learning. It involves dataset preparation, model training, adding SHAP for interpretability, and streamlit implementation.
\subsection{Model Framework for Drug Sensitivity Prediction}
\begin{itemize}
    \item Data Acquisition and Exploration: The dataset acquired from Genomics of Drug Sensitivity in Cancer is loaded into a Pandas DataFrame. The dataset is filtered to only contain LUAD and LUSC cancer types. The target variable, LN-IC50, represents the drug sensitivity (a lower LN-IC50 value indicates higher sensitivity). Using matplotlib, seaborn and plotly, exploratory data analysis is performed to understand the correlation within the dataset.
    \item Data Preprocessing: The missing values of variables greater than 5 percent are dropped and those under are filled in with the model values or mean values. One-hot encoding is performed to convert categorical values into numerical values so that machine learning model can understand.
    \item Model Training: A regressor XGBoost is trained on the selected features with LN-IC50 as target. Hyperparameter search is carried out using Randomized Search. The model is evaluated using RMSE, MAE, and R²  on the test set.
    \item Explainability: SHAP is used to explain which features contribute most to the drug sensitivity prediction for each sample. The SHAP values are then passed on to the DeepSeek API, which summarizes them into simple clinical insights, making the results more interpretable and easy to use in personalized treatment planning.
\end{itemize}  

\subsection{Dataset}
The GDSC (Genomics of Drug Sensitivity in Cancer) dataset represents a complete pharmacogenomic resource aimed at examining the drug response of numerous cancer cell lines. The dataset provides 242,036 records spanning 19 columns of data, directly comprising biological data and molecular [electronic] data. Each cancer cell line is attached to unique identifiers like COSMIC-ID and CELL-LINE-NAME, as is each therapeutic compound like DRUG-ID and DRUG-NAME. The cell lines are denoted with TCGA identifiers (TCGA-DESC, Cancer Type, and Microsatellite instability Status (MSI)). Drug response is quantitatively represented by LN-IC50 (the log-transformed inhibitory concentration),AUC, and Z-SCORE for comparative analyses across drugs and cell lines.
\begin{figure}
    \centering
    \includegraphics[width=1\linewidth]{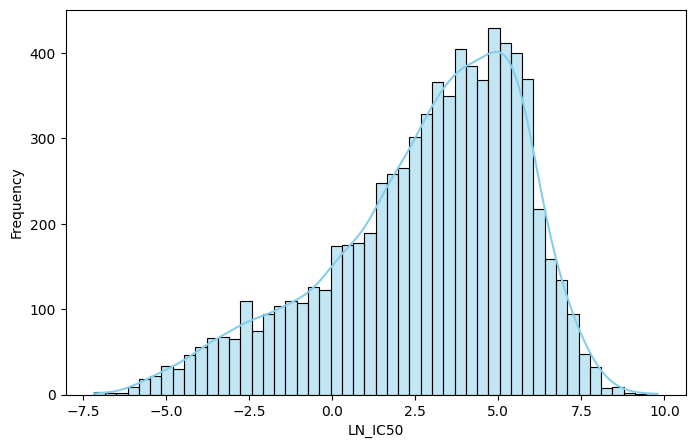}
    \caption{Histogram Showing the Distribution of Log-Transformed IC50 (LN-IC50) Across All Cancer Samples }
    \label{fig:placeholder}
\end{figure}
\begin{figure}
    \centering
    \includegraphics[width=1\linewidth]{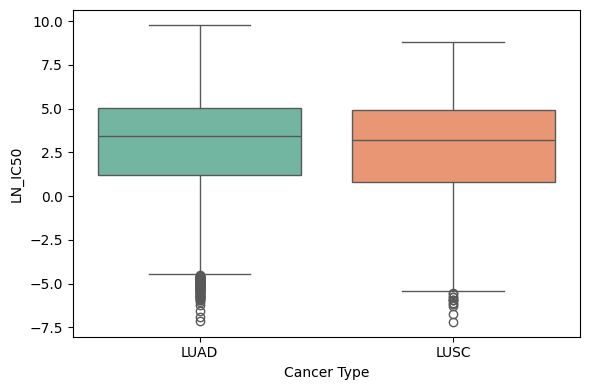}
    \caption{Boxplot representation of LN-IC50 Distribution in LUAD vs LUSC}
    \label{fig:placeholder}
\end{figure}
The dataset provides rich molecular profiling source data also including Gene Expression, Methylation, and CNA (Copy Number Alterations), facilitating research on specific genetic and epigenetic effects on drug sensitivity modeling. The columns TARGET and TARGET-PATHWAY represent the drugs biological targets and affected pathways the drug expresses its action upon. Experimental conditions of screen medium and growth properties in the dataset provide additional contextual information regarding variances in laboratory protocol. The GDSC dataset provides a multidimensional accounting of the biological and molecular properties of multiple cancer types, and avenues for predictive modeling for drug sensitivity in a wide range of clinical circumstances, including identifying how various lung cancer cell types respond to some specific drugs.

\subsection{Data Preprocessing}
To obtain the lung cancer subtypes, the dataset was filtered first to obtain lung cancer subtypes, Lung Adenocarcinoma (LUAD) and Lung Squamous Cell Carcinoma (LUSC) by filtering rows where TCGA-DESC column is either of LUAD or LUSC. Filtering the data in this way helped ensure that the analysis and modeling predictions used the LUAD and LUSC lung cancer subtypes only.
 
Missing data was an important aspect of preprocessing ensuring data quality and accuracy of the model. Any rows with missing values in the TARGET column were removed, as these data would not produce useful information for the model training. The other feature of interest was the Microsatellite instability Status (MSI) feature. For rows without values, the mode (most common value normally) was imputed, as the random imputation would normally introduce bias. Imputing the mode and removing the cells that were null allowed to keep as many values completion as possible in the dataset and not affect the integrity of the MSI feature.Categorical variables were then encoded with one-hot encoding, which transforms categorical columns into numerous binary columns, each of which is categorically 0 or 1 for the specified category. One-hot encoding allows machine learning models to avoid defining ordinal relationships in categorical data while also avoiding assumptions inherent in labeling. Additional features that were not useful for training a model (or were potentially leaking data) were removed from the feature set, including columns of unique identifiers (COSMIC-ID, CELL-LINE-NAME, DRUG-ID), the AUC column, and column values such as labels for cancer type, as the target variable was LN-IC50 (the natural logarithm of the half-maximal inhibitory concentration).  In Figure 2, the histogram represents the distribution of log-transformed IC50 (LN-IC50) values across all cancer cell lines studied. The data have a right-skewed distribution, with most values concentrated between 2.5 and 6.0, which suggests variability of drug sensitivity between samples.

Figure 3 showcases the distribution of LN-IC50 values for two lung cancer subtypes, Lung Adenocarcinoma (LUAD) and Lung Squamous Cell Carcinoma (LUSC). Both subtypes share similar median IC50 values but LUSC potentially slightly greater variability and higher outliers in drug response.

After the dataset is preprocessed, the cleaned dataset was split for training/testing with an 80-20 split to evaluate the model's performance on data it had not previously seen and in order to assess its generalization performance.

\subsection{Development and Implementation of Model}
\begin{figure}
    \centering
    \includegraphics[width=1.2\linewidth]{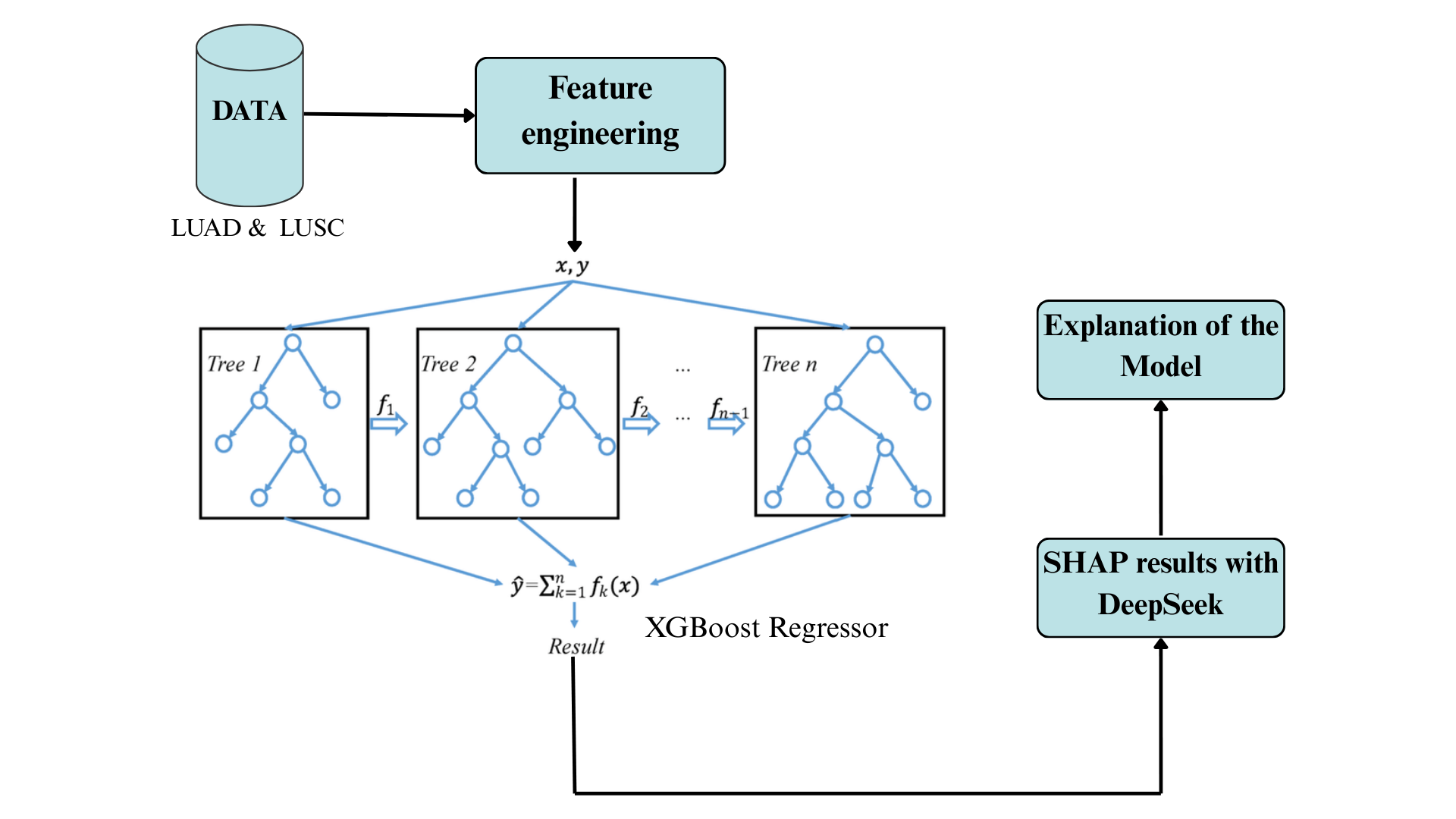}
    \caption{Proposed Predictive Model Architecture for Personalized Lung Cancer Drug Response Treatment }
    \label{fig:placeholder}
\end{figure}

To predict drug sensitivity (LN-IC50), an XGBoost regressor was employed given its good handling of tabular data, and theoretical ability to model complicated functions that have nonlinear non-removable terms. Before fitting the XGBoost model to the training data, a hyperparameter optimization procedure was performed via RandomizedSearchCV. This method searches over a defined parameter space where a random combination of hyperparameters such as the number of trees (n-estimators), learning rate, maximum tree depth (max-depth), subsample ratio, and feature sampling ratio (colsample-bytree) are specified. Multiple combinations of parameter values are evaluated using cross-validation to identify the combination of hyperparameters that yield the best predictive performance based on the R²  metric.
\begin{table}[htbp]
\centering
\renewcommand{\arraystretch}{1.15} 
\setlength{\tabcolsep}{8pt}        
\normalsize 

\caption{Regression Model Performance Comparison}
\label{tab:model_performance}

\begin{tabular}{lccc}
\hline
\hline
\textbf{Model} & \textbf{MAE} & \textbf{MSE} & \textbf{R\textsuperscript{2}} \\
\hline
XGBoost                   & 0.0851 & 0.0249 & 0.9971 \\
Random Forest             & 0.8228 & 1.1032 & 0.8700 \\
Linear Regression         & 0.2268 & 0.1298 & 0.9847 \\
\hline
\end{tabular}
\end{table}

Once the optimal hyperparameters were selected, we then fit the XGBoost model on the training data. The fitted model was then evaluated on the hold-out test set to assess performance using regression error metrics, Mean Squared Error (MSE), Mean Absolute Error (MAE) and R²  score. This quantified the amount of prediction error, derived some insights into the model's performance, and informed us about the amount of variability explained in drug sensitivity. To enhance the strength and reliability of the model, a process of 10-fold cross-validation was performed. In this process, the entire dataset was divided into 10 subsets (folds), and the model was trained on 9 out of the 10 folds while validation was done on the remaining fold. This process was repeated 10 times. This not only helped evaluate the reliability of the model on certain subsets of data, but it also helped to control overfitting and gave a better generalized estimate of model performance.
\vspace{1em}
\rule{\linewidth}{0.4pt}
\textbf{Algorithm 1} Drug Sensitivity Prediction and Explainability for Lung Cancer Treatment\\
\rule{\linewidth}{0.4pt}
\textbf{Input:} GDSC Dataset (filtered for LUAD and LUSC)\\
\textbf{Output:} LN\_IC50 prediction model with SHAP-based clinical recommendations
\begin{algorithmic}[1]

\STATE \textbf{Step 1: Load and Filter Data}
\STATE Load the GDSC dataset into a dataframe
\STATE Filter rows where \texttt{TCGA\_DESC} $\in$ \{\texttt{LUAD, LUSC}\}

\STATE \textbf{Step 2: Handle Missing Values}
\FOR{each column in the dataset}
  \IF{column is \texttt{TARGET} and value is missing}
    \STATE Drop the row
  \ELSIF{column is \texttt{MSI Status}}
    \STATE Fill missing values with most frequent value (mode)
  \ENDIF
\ENDFOR

\STATE \textbf{Step 3: Feature Selection and Encoding}
\STATE Drop identifier columns
\STATE Set target variable $y \leftarrow$ \texttt{LN\_IC50}
\STATE One-hot encode categorical features in $X$

\STATE \textbf{Step 4: Train-Test Split}
\STATE Split dataset into 80\% training and 20\% testing sets

\STATE \textbf{Step 5: Hyperparameter Tuning}
\STATE Define search space for \texttt{XGBRegressor}
\STATE Perform \texttt{RandomizedSearchCV} with 5-fold cross-validation

\STATE \textbf{Step 6: Model Training and Evaluation}
\STATE Train final model using best parameters
\STATE Predict LN\_IC50 on test data
\STATE Compute R\textsuperscript{2} score and MAE
\STATE Evaluate with 10-fold cross-validation

\STATE \textbf{Step 7: Explainability with SHAP}
\STATE Initialize SHAP TreeExplainer with trained model
\STATE Compute SHAP values on test set
\STATE Plot SHAP summary and bar charts
\STATE Rank features by average SHAP importance

\STATE \textbf{Step 8: Instance-Level Interpretation}
\STATE Select test instance $i$ with index $true\_index$.
\STATE Retrieve predicted LN\_IC50 and associated \texttt{DRUG\_NAME}.
\STATE Determine response type: resistant if $ LN\_IC50 > 4$, sensitive otherwise
\STATE Extract top 5 contributing SHAP features

\STATE \textbf{Step 9: Clinical Summary Generation}
\STATE Construct prompt using SHAP features and prediction
\STATE Send prompt to DeepSeek API for interpretation
\IF{API call is successful}
  \STATE Print clinical explanation and actionable insights
\ELSE
  \STATE Log API error or retry
\ENDIF

\STATE \textbf{Step 10: Streamlit App Deployment }
\STATE Design interface for input features and visualization
\STATE Display model prediction, SHAP plots, and DeepSeek output

\end{algorithmic}

\rule{\linewidth}{0.4pt}
\begin{figure}
    \centering
    \includegraphics[width=1\linewidth]{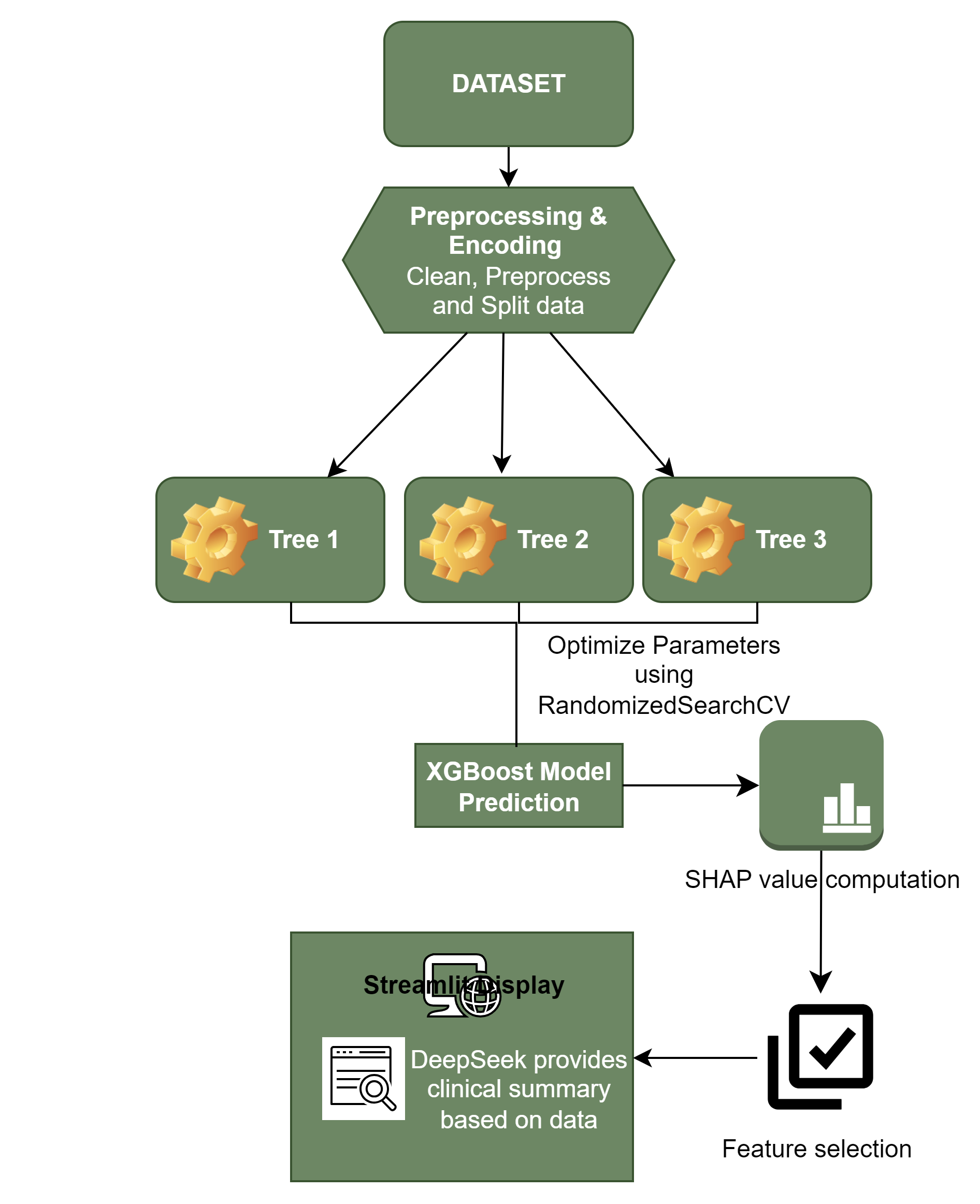}
    \caption{Model Flowchart for Drug Response Prediction System}
    \label{fig:placeholder}
\end{figure}
\subsection{Explainability}
SHapley Additive exPlanations (SHAP) was used in this case to interpret the predictions generated by the XGBoost model to predict drug response. SHAP is a game-theoretic technique which allocates a contribution value to each feature and describes the change in any individual prediction. SHAP can provide detailed interpretability globally or locally, and provides a richer description of model behavior than merely looking at feature importance, as we can describe the impact of features on a given model output. 

For the XGBoost model,  TreeExplainer method was used specifically designed to calculate exact SHAP values for all the features for the entire test dataset using tree-based models such as XGBoost. Summary plots of SHAP values-once we assigned values using the scanner procedure-allowed us to determine which features had the largest overall impact on the predicted drug sensitivity. A mean absolute SHAP value bar plot provided further clarity about which features had the largest absolute effect on the model's predicted drug response. 

Local interpretability was illustrated by reviewing the SHAP values for each individual test sample. For each patient, the features were ranked based on the size of the SHAP values, and identified the top genomic or clinical features that most influenced the predicted drug response (LN-IC50) as shown in Figure 6. This personalized interpretation is extremely valuable for precision medicine as it reveals patient specific drivers of drug sensitivity or resistance. To increase clinical relevance, the SHAP-based explanations were bundled into a clinical assistant AI via the DeepSeek API. The clinical AI was provided with information on the predicted drug response, drug name, and the top contributing features. It then generated a report conveying a complete interpretation including the mechanism of action of the drug, and considerations for drug metabolism along with personalized treatment recommendations. This improved the transparency in the use of AI, easing the translation of complicated outputs from machine learning models into applicable clinical behaviour.

In summary, the SHAP framework provides a complete way for explaining predictions in this regard. It combines both global information concerning which features were the most important predictors for the cohort, as well as individual explanations to aid personalized recommendations. This work contributes toward an effective and safe implementation of AI in precision oncology.

\begin{figure}
    \centering
    \includegraphics[width=1\linewidth]{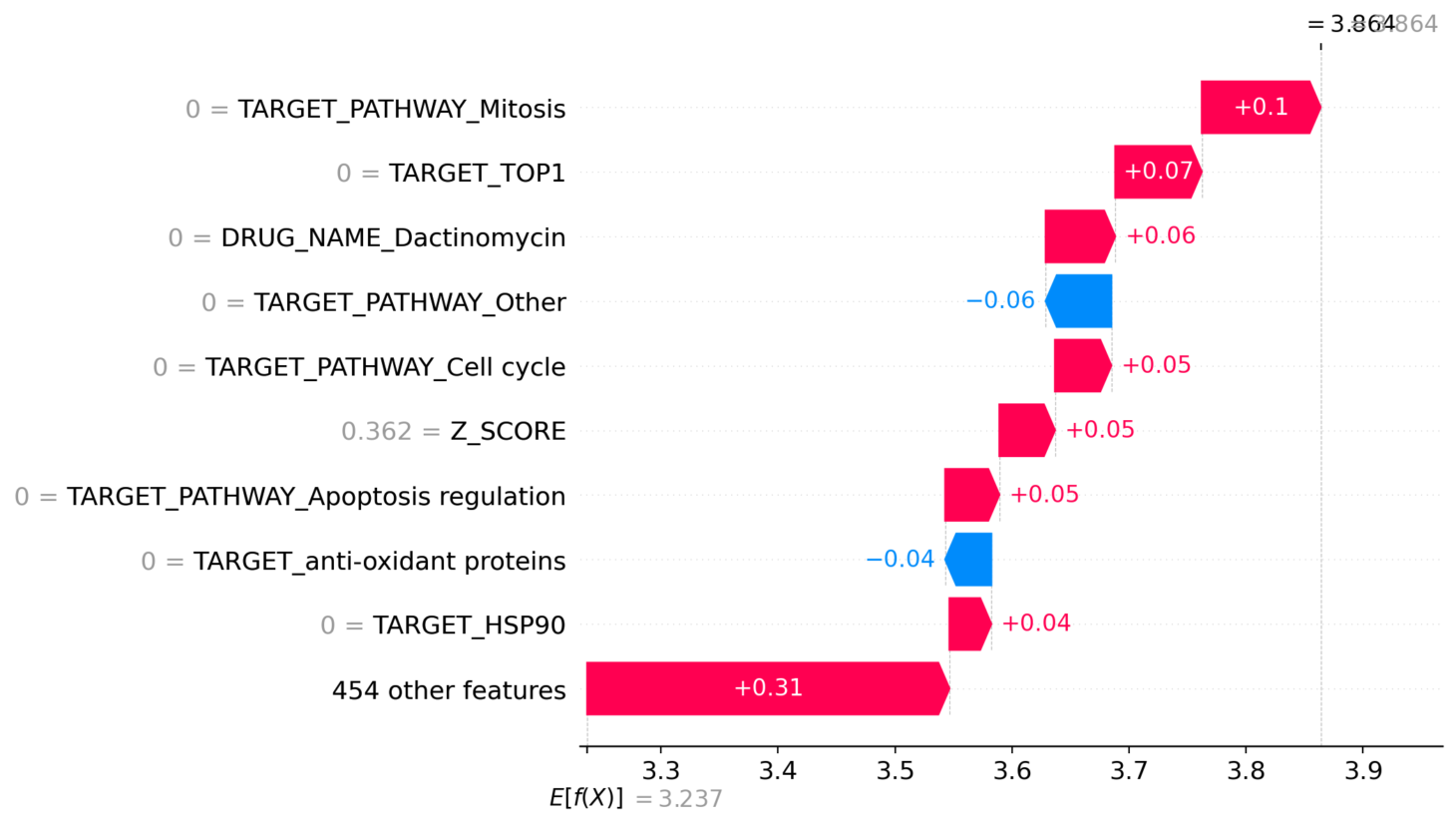}
    \caption{SHAP Waterfall Plot: Contribution of each Feature}
    \label{fig:placeholder}
\end{figure}

\begin{figure}
    \centering
    \includegraphics[width=1\linewidth]{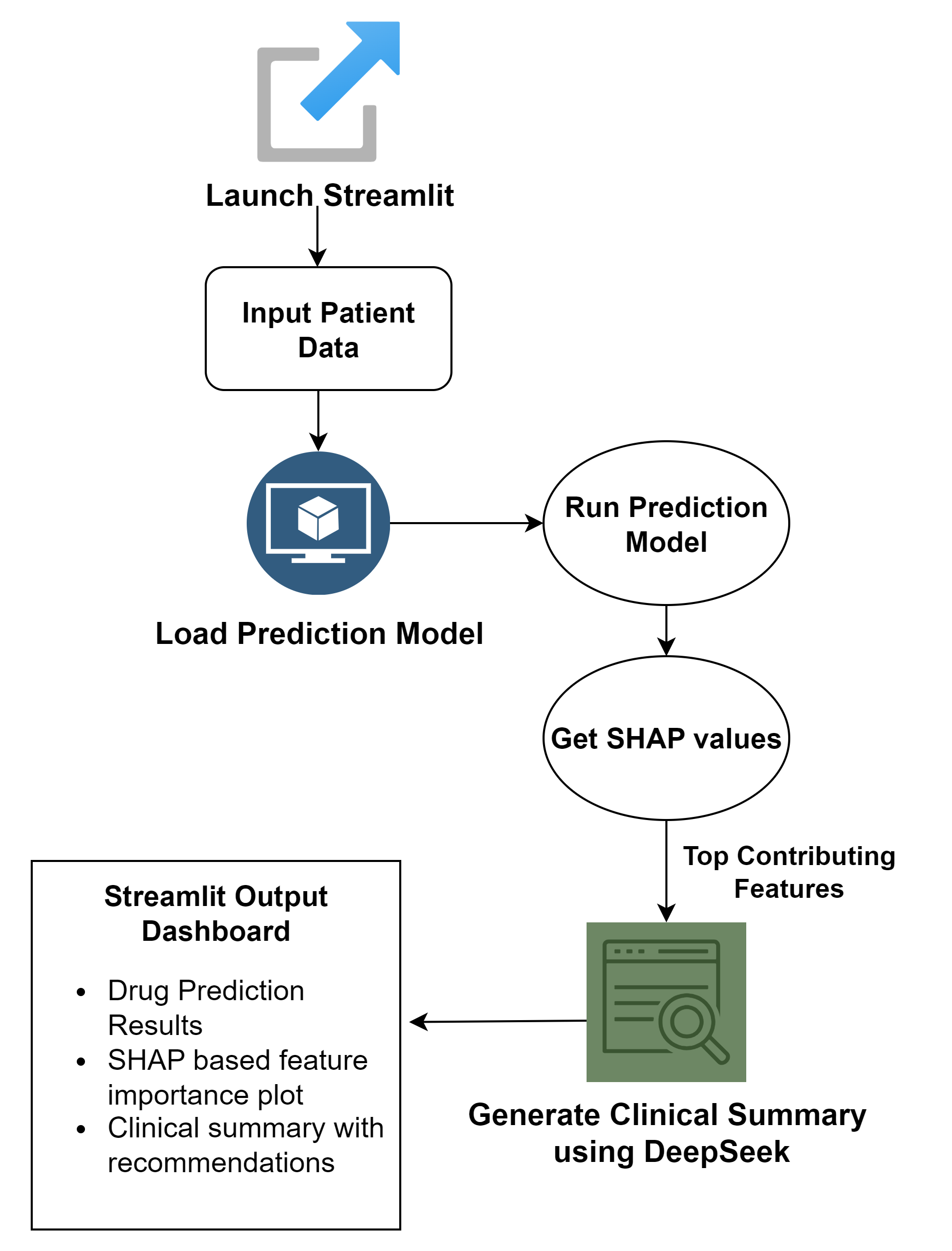}
    \caption{Workflow diagram of the end-user application developed to make use of the proposed model}
    \label{fig:placeholder}
\end{figure}
\section{Results and Discussion}

A relevant study by Pant et al. [13] introduced a machine learning framework to predict cancer drug sensitivity using genomic features. This study used the Genomics of Drug Sensitivity in Cancer (GDSC) dataset and performed several preprocessing steps such as feature engineering, dimensionality reduction through Principal Component Analysis (PCA) and autoencoders, and employed a variety of modeling strategies. The study explored logistic regression, random forest and XGBoost; however the random forest model was the most robust for average validation metrics. The study reported a coefficient of determination (R²  ) of 0.99, a mean absolute error (MAE) of 0.16, mean squared error (MSE) of 0.08 and a root mean squared error (RMSE) of 0.29.

In contrast the current methodology only analyzes LUAD and LUSC subtypes in the GDSC and reframes the analysis from binary classification to regression problems predicting LN-IC50 values. This reframing allows the model to examine more nuanced levels of drug sensitivity. Each approach is trained on an optimized XGBoost regressor with hyperparameter tuning from RandomizedSearchCV. As a result of the new modeling approach, the final model had improved predictive performance including R²  of 0.9971, MAE of 0.0851 and MSE of 0.0249 indicating improved accuracy and generalization. In addition to predictive accuracy, the method focuses on explainability. This is not a new aspect of machine learning in precision medicine, but the previously reported study did mention challenges with interpretability of high-dimensional models relative to the future clinical application of models. The current method addresses this through the SHAP (SHapley Additive exPlanations) algorithm to offer global and local feature attribution. This can be further enhanced through the DeepSeek API consumer natural language clinical interpretations based on the list of top contributors features for each prediction. This provides a more clinically relevant interpretation of the complex predictions. 

Also, the framework is deployed through a Streamlit based application to provide end users the ability to input data interactively, view predictions, visualize SHAP outputs, and view clinical summaries. The end-to-end pipeline from data preprocessing and regression modeling through explainability to real-time deployment create a holistic and interpretable approach to advance personalized treatments of lung cancer based on genomic profiles.

\begin{table}[ht]
\caption{Comparison of Model Performance Metrics}
\centering
\begin{tabular}{|l|c|c|}
\hline
\textbf{Metric} & \textbf{Previous Study [13]}& \textbf{Current Work (XGBoost + SHAP)} \\
\hline
R\textsuperscript{2} Score & 0.9900 & 0.9971 \\
\hline
MAE & 0.1600 & 0.0851 \\
\hline
MSE & 0.0800 & 0.0249 \\
\hline
RMSE & 0.2900 & 0.1578 \\
\hline
\end{tabular}
\label{tab:comparison}
\end{table}

\label{sec:guidelines}

\begin{figure}
    \centering
    \includegraphics[width=1\linewidth]{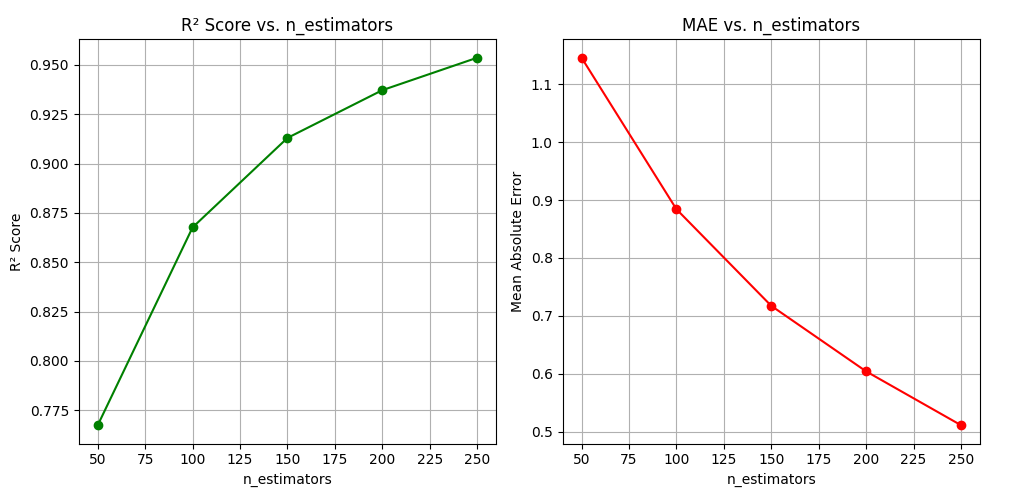}
    \caption{Model Performance with Varying Boosting Rounds}
    \label{fig:placeholder}
\end{figure}

To simulate the effect of shifting epochs of training, model performance is assessed by different values of n-estimators (XGBoost iterations). As shown in Figure 8, R²  score progressively increased with rising estimators from 0.77 (50 estimators) to 0.95 (250). Also, MAE decreased from 1.15 to 0.51, indicating greater accuracy of prediction.
\section{Model Performance and Overfitting Assessment}

The model's performance was evaluated using 5-fold cross-validation to assess its generalization capability. The results are summarized in Table~\ref{tab:cv-results}.

\begin{table}[h!]
\centering
\caption{Cross-Validation Performance Metrics}
\label{tab:cv-results}
\begin{tabular}{lcccccc}
\toprule
Metric & Fold 1 & Fold 2 & Fold 3 & Fold 4 & Fold 5 & Average\\
\midrule
R²  Score & 0.9979 & 0.9967 & 0.9944 & 0.9974 & 0.9963 & 0.9965 \\
MAE      & 0.0746 & 0.0755 & 0.0884 & 0.0786 & 0.0798 & 0.0794 \\
\bottomrule
\end{tabular}
\end{table}
5-fold Cross validation was used to determine the performance of the model and test its ability to generalize. The R²  scores achieved on the folds were: 0.9979, 0.9967, 0.9944, 0.9974 and 0.9963, which gave a mean of 0.9965. The corresponding Mean Absolute Error (MAE) values were 0.0746, 0.0755, 0.0884, 0.0786 and 0.0798 with the average MAE of 0.0794. The high R²  values nearly equal to 1.0, and the low values of the MAE in all folds point to the fact that the model has high predictive accuracy. The low difference in cross-validation folds also shows consistency of performance across subsets of data.

These findings are an indication that the model is not over-fitting the training set. Differences in performance between folds would be quite large or a large decline of accuracy on unseen data would be observed to characterize overfitting. This has been observed to be consistent and therefore implies that the model generalizes well to unseen new samples because of capture of patterns in data, not memorization of noise.

\section{Conclusion}
 In this paper, a machine learning framework is presented to predict drug responsiveness in lung cancer subtypes LUAD and LUSC, using the Genomics of Drug Sensitivity in Cancer (GDSC) dataset . By developing an XGBoost regression model and down-sampling the hyperparameter optimization via a randomized tuning process, the model achieved an R²  score of 0.9971 and very low error, which outperformed the regression model Random Forest and Linear Regression. In addition, SHAP
 values was utilized to interpret the model because it allowed to measure important genomic and clinical features that drove drug sensitivity and resistance. Interpretable models are especially important for understanding complex biological interactions, and support decision-making in precision medicine. Additionally, DeepSeek API was used to summarize clinical observations and link it back to the web-based drug predictions, highlighting the practical implications of using explainable AI methods for precision oncology applications. The significance of this work is that it can ultimately help improve therapeutic success rates by tailoring drug regimens based on an individual’s molecular profile and eliminate the burden of trial and error, which is often the case in cancer drug decisions. Potential uses of this framework includes helping clinicians figure out which drugs to try first, identifying which resistance mechanism might occur, and identifying worthy candidate biomarkers for further exploration in the area of precision oncology.

\section{Declarations}
\begin{itemize}
    \item Availability of data and material: Not applicable
    \item Competing interests: Not applicable
    \item Funding: Not applicable
    \item Authors' contributions: Ann Rachel directed the study's conception and design, conducted a comprehensive literature review, structured the taxonomy of challenges and strategies, and authored the initial draft of the manuscript. Pranav M. Pawar contributed to the drafting and enhancement of key sections of the manuscript, oversaw the research process, and participated in the analysis of digital privacy risks associated with IoT, AI/ML, and federated learning solutions. Mithun Mukherjee engaged in manuscript revision, facilitated the critical assessment of existing methodologies, and contributed to discussions regarding unresolved issues and prospective research directions. Raja M offered essential feedback, assessed and refined the manuscript for technical rigor and clarity, and guided the research trajectory. Tojo Mathew assessed and refined the manuscript for technical rigor and clarity, and guided the research trajectory
    \item Acknowledgements: Not applicable
\end{itemize}



\begin{thebibliography}{}
%
%
\bibitem{b1} World Health Organization, "Lung cancer," WHO, Fact Sheets, 2023. [Online]. Available: https://www.who.int/news-room/fact-sheets/detail/lung-cancer.

\bibitem{b2}M. S. Abrazinski, "Lung Cancer Overview," Medscape, 2024. [Online]. Available: https://emedicine.medscape.com/article/279960-overview. 

\bibitem{b3} J. Smith et al., "Advanced Imaging in Lung Cancer Diagnosis," The British Journal of Radiology, vol. 96, no. 1152, p. 20230334, 2023. [Online]. Available: https://academic.oup.com/bjr/article/96/1152/20230334/7499290.

\bibitem{b4}A. P. Smith, B. Jones, and C. Taylor, "Lung Cancer Treatment and Progress," PMC, 2019. [Online]. Available: https://pmc.ncbi.nlm.nih.gov/articles/PMC6352312/. 


\bibitem{b5}D. Stefanicka-Wojtas and D. Kurpas, “Personalised Medicine—Implementation to the Healthcare System in Europe (Focus Group Discussions),” Journal of Personalized Medicine, vol. 13, no. 3, p. 380, Feb. 2023, doi: https://doi.org/10.3390/jpm13030380.

\bibitem{b6}P. Krzyszczyk et al., “The growing role of precision and personalized medicine for cancer treatment,” TECHNOLOGY, vol. 06, no. 03n04, pp. 79–100, Sep. 2018, doi: https://doi.org/10.1142/s2339547818300020.

\bibitem{b7} Deshmukh SK. , "Artificial Intelligence and Machine Learning in Cancer Care: Current Applications and Future Perspectives,” \textit{Journal of Cancer Immunology}, vol. 2, no. 2, Aug. 2020, doi: https://doi.org/10.33696/cancerimmunol.2.011.‌
\bibitem{b8} Chen \textit{et al.}, “Integrative Machine Learning Approach for Forecasting Lung Cancer Chemosensitivity: From Algorithm to Cell Line Validation,” \textit{Computational and Structural Biotechnology Journal}, vol. 27, pp. 3307–3318, Jan. 2025, doi: https://doi.org/10.1016/j.csbj.2025.07.043.‌

\bibitem{b9} A. Hooshmand, “Machine Learning Against Cancer: Accurate Diagnosis of
Cancer by Machine Learning Classification of the Whole Genome Sequencing Data,” arXiv.org, Sep. 12, 2020.https://arxiv.org/abs/2009.05847

\bibitem{b10} D. Oniani et al., “Comparisons of Graph Neural Networks on Cancer Classification Leveraging a Joint of Phenotypic and Genetic Features,” arXiv, Jan. 2021.
https://doi.org/10.48550/arxiv.2101.05866

\bibitem{b11} B. Bhinder et al., “Artificial Intelligence in Cancer Research
and Precision Medicine,” Cancer Discovery, vol. 11, no. 4,2021. https://doi.org/10.1158/2159-8290.cd-21-0090

\bibitem{b12} N. Zong et al., “Leveraging Genetic Reports and Electronic Health Records
for the Prediction of Primary Cancers,” JMIR Med. Inform., vol. 9, no. 5, 2021.
https://doi.org/10.2196/23586

\bibitem{b13} “Index - FHIR v4.0.1,” \textit{www.hl7.org}. https://www.hl7.org/fhir/ 

\bibitem{b14} S. D’Amico et al., “Synthetic Data Generation By Artificial Intelligence to
Accelerate Translational Research and Precision Medicine in Hematological Malignancies,” Blood, vol. 140, no. Suppl. 1, pp. 9744–9746, Nov. 2022. https://doi.org/10.1182/blood-2022-
168646

\bibitem{b15} S. J. MacEachern and N. D. Forkert, “Machine learning for precision medicine,” Genome, vol. 64, no. 4, pp. 416–425, Apr. 2021, doi: https://doi.org/10.1139/gen-2020-0131.

\bibitem{b16}B. Ghoshal and A. Tucker, “Leveraging Uncertainty in Deep Learning for Pancreatic Adenocarcinoma Grading,” arXiv (Cornell University), 2022, doi: https://doi.org/10.48550/arxiv.2206.08787

\bibitem{b17} A. Rafferty, R. Ramaesh, and A. Rajan, “Transparent and Clinically
Interpretable AI for Lung Cancer Detection in Chest X-Rays,” arXiv (Cornell University), Mar. 2024, https://doi.org/10.48550/arxiv.2403.19444

\bibitem{b18} Y. Jiang, M. Chen, Z. Xiong, and Y. Qin, "Predicting anti-cancer drug
sensitivity through WRE-XGBoost algorithm with weighted feature selection," Genes and Diseases, vol. 12, no. 2, p. 101275, Mar. 2024, doi:10.1016/j.gendis.2024.101275

\bibitem{b19} R. Qureshi, S. A. Basit, J. A. Shamsi, X. Fan, M. Nawaz, H. Yan,
and T. Alam, "Machine learning based personalized drug response prediction for lung cancer patients," Scientific Reports, vol. 12, no. 1, p. 18935, Nov.2022, doi: 10.1038/s41598-022-23649-0

\bibitem{b20} J. R. Astley, J. M. Reilly, S. Robinson, J. M. Wild, M. Q. Hatton, and B. A. Tahir, “Explainable deep learning-based survival prediction for non-small cell lung cancer patients undergoing radical radiotherapy,” Radiotherapy and Oncology, vol. 193, pp. 110084–110084, Jan. 2024, doi: https://doi.org/10.1016/j.radonc.2024.110084

\bibitem{b21} ]L. Pant, A. A. Mukaddim, M. K. Rahman, A. A. Sayeed, M. S. Hossain,
M. T. Khan, and A. Ahmed, "Genomic predictors of drug sensitivity in cancer:
Integrating genomic data for personalized medicine in the USA," vol. 5, no. 12, Dec. 2024.

\bibitem{b22} ]Y. Shi, C. Li, X. Zhang, C. Peng, P. Sun, Q. Zhang, L. Wu, Y. Ding, D.
Xie, Z. Xu, and W. Zhu, “D3EGFR: a webserver for deep learning-guided
drug sensitivity prediction and drug response information retrieval for EGFR mutation-driven lung cancer,” Briefings in Bioinformatics, vol. 25, no. 3, Mar. 2024, Art. no. bbae121, doi:10.1093/bib/bbae121.
\end{thebibliography}


\end{document}